%% file: main.tex
\documentclass{article}

\PassOptionsToPackage{numbers, compress}{natbib}
 \usepackage[dblblindworkshop,final]{neurips_2025}
\usepackage[utf8]{inputenc} % allow utf-8 input
\usepackage[T1]{fontenc}    % use 8-bit T1 fonts
\usepackage{hyperref}       % hyperlinks
\usepackage{url}            % simple URL typesetting
\usepackage{booktabs}       % professional-quality tables
\usepackage{amsfonts}       % blackboard math symbols
\usepackage{nicefrac}       % compact symbols for 1/2, etc.
\usepackage{microtype}      % microtypography
\usepackage{xcolor}         % colors

\usepackage{graphicx}\usepackage{subcaption}
\usepackage{booktabs}
\usepackage{siunitx}
\usepackage{adjustbox}

\newcommand{\custompar}[1]{\noindent\textbf{#1:\;}}

\definecolor{mynangrey}{rgb}{0.85, 0.85, 0.85}

\title{TabGemma: Text-Based Tabular ICL via LLM using Continued Pretraining and Retrieval}
\workshoptitle{AI for Tabular Data}
\author{%
Günther Schindler\quad%
Maximilian Schambach\quad%
Michael Medek\quad%
Sam Thelin\\
SAP SE\\
\texttt{\{firstname.lastname\}@sap.com}
}

\begin{document}

\maketitle

\begin{abstract}
We study LLMs for tabular prediction with mixed text, numeric, and categorical fields. We introduce TabGemma, a schema-agnostic in-context learner that treats rows as sequences and tackles two practical hurdles when adapting pretrained LLMs for tabular predictions: unstable numeric tokenization and limited context size. We propose to canonicalize numbers via signed scientific notation and continue pretraining of a 12B Gemma 3 model with a target imputation objective using a large-scale real world dataset. For inference, we use a compact n‑gram–based retrieval to select informative exemplars that fit within a 128k-token window.

On semantically rich benchmarks, TabGemma establishes a new state of the art on classification across low- and high-data regimes and improves monotonically with more context rows. For regression, it is competitive at small sample sizes but trails conventional approaches as data grows. Our results show that LLMs can be effective tabular in-context learners on highly semantic tasks when paired with dedicated numeric handling and context retrieval, while motivating further advances in numeric modeling and long-context scaling.
\end{abstract}

\section{Introduction}
Many real-world tabular prediction tasks include rich textual information such as descriptive column headers, semantically meaningful categoricals or free-text columns alongside numeric and date-like features. Classical tabular predictive models such as gradient-boosted trees excel on structured inputs but typically lack fine-grained semantic understanding of such text. In practice, this gap is bridged with hand-crafted features, bag-of-words or TF-IDF vectors, or separate text encoders glued to tabular pipelines, all of which add complexity and reduce portability across schemas and domains~\cite{grinsztajn2023vectorizing}. 

While recent advances in tabular deep learning achieved impressive performance via end-to-end trained in-context learning (ICL), outperforming conventional approaches in some domains, as pioneered by TabPFN~\cite{tabpfnv1, tabpfnv2} and extended in other works~\citep{tabicl, tabdpt}, most of these methods also do not make explicit use of the semantic content within the data and rely on conventional feature encodings. Only the recent ConTextTab approach integrates semantic embeddings into a table-native ICL architecture~\cite{contexttab} but compresses potentially large free-text cells into a single embedding vector, potentially limiting its semantic expressivity at scale.

Large language models (LLMs) offer a compelling alternative: they can consume heterogeneous data types by serializing tables as text and perform classification or regression via in-context learning, bringing strong semantic capabilities to the textual fields while handling mixed data types through a single interface. However, two hurdles hinder practical deployment for tabular prediction: raw decimals tokenize poorly and inconsistently~\cite{numeric_tokenizer}, and the finite context window limits how many relevant exemplars can be provided as context, especially in large datasets. In the past, this has severly limited performance of LLM-based tabular predictors.

Building on prior work that serializes tables for LLMs and using retrieval-augmented generation to scale ICL, we present TabGemma, a schema-agnostic method that treats tabular prediction as sequence modeling, improves numeric tokenization, and selects informative exemplars via efficient retrieval at inference. Rows are serialized with dedicated separators and numerics are canonicalized into signed base-10 scientific notation, yielding stable token patterns. We continue pretraining of a 12B Gemma~3~\cite{gemma3} model on a column-imputation objective that applies loss only to target-column tokens while conditioning on all feature tokens, aligning next-token prediction with classification/regression. To fit task-relevant support within the context window at inference, we perform nearest-neighbour retrieval using compact hashed character n-gram embeddings per cell, concatenated into row embeddings and indexed with FAISS~\cite{faiss}: at inference, we retrieve k similar rows, serialize them as exemplars, and append the query row with an empty target for the model to decode.

\section{Related Work}
\custompar{Tabular deep learning} Prediction on tabular data has long been dominated by boosted trees such as XGBoost, LightGBM, and CatBoost~\cite{xgboost, lightgbm, catboost}. While strong, these models need to be trained per dataset, cannot benefit from cross‑task pretraining, and often require manual feature engineering and extensive hyperparameter optimization. Early deep learning architectures like FT‑Transformer~\cite{ft-transformer} and XTab~\cite{xtab} explored transformer‑based encoders, while only more recent methods, e.g.\ TabR, RealMLP, CARTE, TabM, or ModernNCA~\cite{tabr, realmlp, CARTE, tabm, modernnca}, report consistently competitive, sometimes superior results to boosted trees.

\custompar{In‑context learning for tabular data} TabPFNv1~\cite{tabpfnv1} demonstrated that row‑level ICL pretrained on synthetic tasks can outperform boosted trees on small classification problems, eliminating per‑task training and hyperparameter tuning. Using real data and retrieval to select context examples, TabDPT achieved similarly strong results and extended the setting to regression, building on ideas also investigated in TabR. Moving beyond row‑level encodings, cell‑based ICL, as used in TabPFNv2~\cite{tabpfnv2}, TabICL~\cite{tabicl}, and ConTextTab~\cite{contexttab} scale to larger datasets and report state-of-the-art results.

\custompar{Semantics and real data} Capturing fine‑grained semantics in real‑world tables is key for transfer beyond statistical patterns. CARTE~\cite{CARTE} pretrains across diverse sources to model table semantics and achieves state of the art on its benchmark, but requires task‑specific fine‑tuning. Modern LLMs bring stronger semantic understanding and world knowledge but lack native table support. Several works adapt LLMs to tabular ICL, for example TabLLM, LIFT, and TabuLa‑8B~\cite{tabllm, lift, tabula8b}, which also curates the T4 dataset -- a collection of 3M tables derived from TabLib~\cite{tablib} -- and show excellent performance in the very low‑data regimes.

\section{Methodology}
\begin{figure}
\centering 
\includegraphics[width=\linewidth]{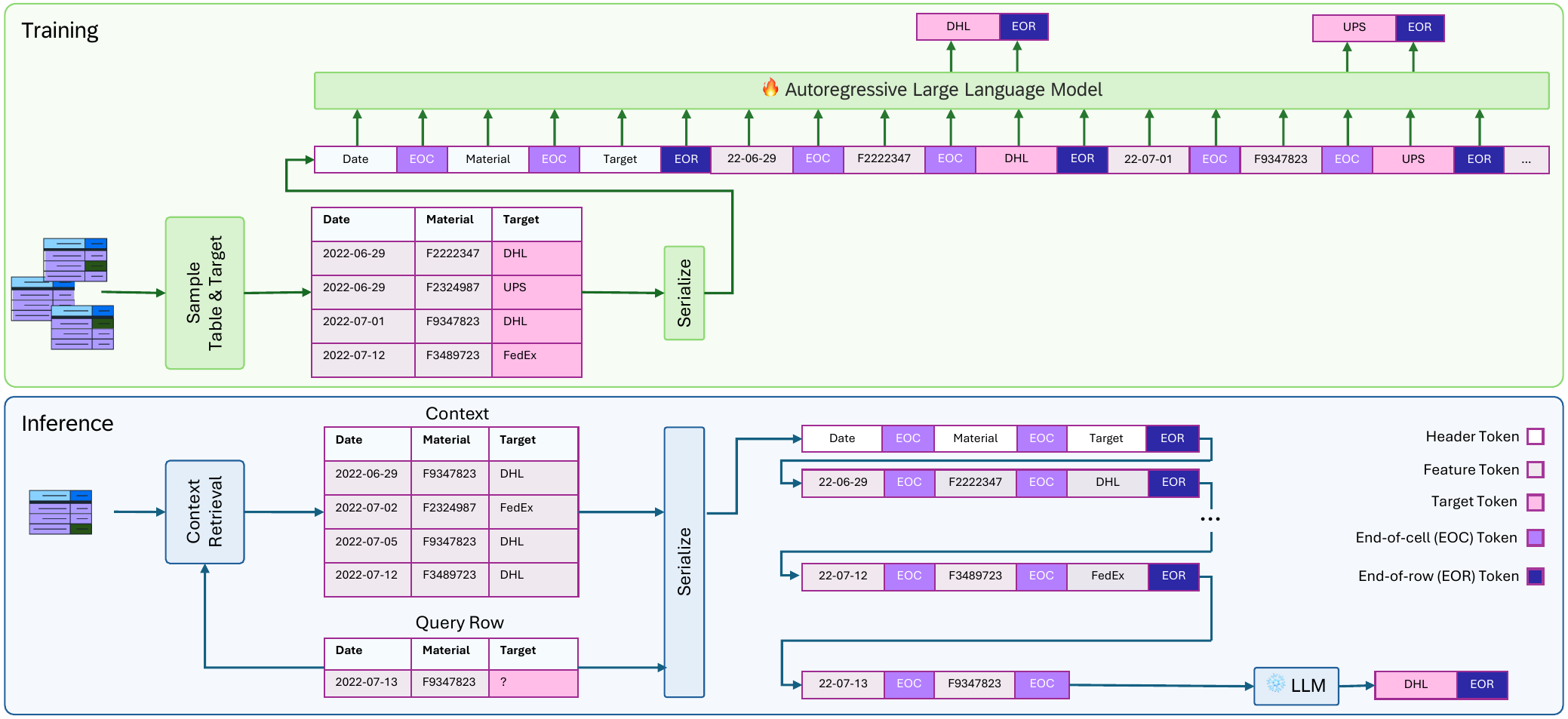}
\caption{Illustration of our proposed LLM-based tabular prediction architecture with table serialization and target-imputation objective at training, and local context retrieval at inference.} 
\label{fig:method-overview} 
\end{figure}

Similar to TabuLa, we cast tabular classification and regression as sequence modeling: the input table is serialized into tokens, and a long‑context LLM is trained to predict a designated target column causally conditioned on feature tokens. The method is schema‑agnostic and centers on three components: canonical row serialization, continued pre‑training with a target‑imputation objective, and similarity-based retrieval to fit task‑relevant exemplars within the context window at inference. An overview of our proposed approach is depicted in Figure~\ref{fig:method-overview}.

\custompar{Table serialization}
We serialize each table row into a linear token sequence. Every cell is first cast to a canonical string, tokenized, and concatenated in column order. Cells are separated by a dedicated cell-separator token. Each row is terminated with an end-of-row token, which conditions the model to stop decoding once the target has been produced. The input to the model is the concatenation of such row sequences. In particular, we do \emph{not} prepend a task-specific natural-language instruction for the LLM.
Numeric values are normalized before tokenization using signed, base-10 scientific notation with four significant digits. For example, 3141.592 becomes +3.1416e+03. This canonicalization is locale-independent and reduces variability. In subword tokenizers, scientific notation induces reusable subpatterns (e.g., "+", "e+0"), which improves the learning of token-based numeric embeddings compared to raw decimals, whose lengths and delimiters vary widely.

\custompar{Continued pretraining}
We initialize from the pretrained Gemma 3 12B checkpoint (without instruction‑tuning), which supports a 128k‑token context window, and continue pretraining the model on a tabular imputation objective tailored to classification and regression over tables using the large-scale T4 table corpus~\cite{tabula8b}.
For each training step, we draw a table from the corpus, uniformly sample 256 rows, and designate one column as the prediction target while using the remaining columns as conditioning features. The input consists of all selected rows with their full feature columns and the target column values present in the text for teacher forcing. We compute token‑level cross‑entropy only over the tokens belonging to the target column, masking out loss contributions from feature columns. The prompt format scales naturally to different numbers of rows at inference time, allowing the model to condition on variable‑sized contexts. Note that, due to the autoregressive nature of LLMs and the use of causal attention masking, \emph{every} target cell is used as a training sample and the approach does not require a fixed context-query split unlike table-native approaches such as TabPFN -- practically increasing the effective batch size as well as conditioning the model on different effective context lengths within a single training step. 

\custompar{Inference \& Retrieval}
Scaling in-context learning to larger tables is constrained by the model’s context window. As compared to table-native architectures, this becomes even more pronounced when adapting LLMs for tabular ICL due to the relatively less efficient tokenization scheme. To mitigate this limitation, we employ similarity-based context retrieval to select a small, task-relevant support set for each query at inference. Our approach is similar to that of TabR and TabDPT but note that we use it during inference only while we found it sufficient to randomly sample context during training. We use nearest-neighbour retrieval by constructing compact row embeddings via hashing: Each table row is serialized as a sequence of cells, and each cell is independently vectorized using a bag of character n-grams. The n-grams are hashed into a fixed 256-dimensional vector per cell. The per-cell vectors are concatenated to form the row embedding. This representation is stateless, memory-efficient, and offers parallelization over rows, enabling streaming ingestion and scaling to millions of training rows. For similarity search, we L2-normalize embeddings and index them with FAISS~\cite{faiss}. The retrieved rows are then formatted as ICL exemplars and provided to the LLM.

\section{Experiments}
\custompar{Training} We train for 2,500 steps with a batch size of 64, corresponding to roughly 41 million row predictions in total. We uses Adam with a learning rate of $10^{-5}$. We do not apply dropout or weight decay. To balance throughput and context utilization, we cap inputs at 16k tokens during training and truncate sequences that exceed this limit. We trained the model for ~20 days on 2 H100 GPUs.

\custompar{Evaluation} We evaluate our approach on several benchmarks: the CARTE benchmark~\cite{CARTE}, the recently proposed TextTab benchmark~\cite{texttabbench}, as well as the recent TabArena benchmark~\cite{tabarena} in its ``lite'' variant (evaluating a single fold).
All benchmarks are mixed classification and regression benchmarks.
Whereas CARTE and TextTab are constructed to emphasize semantic features and are thus our focus, TabArena is a more conventional, numerics-heavy benchmark which we include for completeness.

We compare against a range of extensively tuned conventional as well as deep learning and ICL baselines, including LGBM, RealMLP, TabPFN, ConTextTab, Random Forest, and a naive predictor. Additionally, we include results of the AutoML framework AutoGluon~\cite{autogluon}. The details about the used baselines can be found in Appendix~\ref{app:baselines}. Unfortunately, despite much effort spent, we were not able to run LLM-based baselines such as TabuLa~\cite{tabula8b} or GTL~\cite{gtl} due to non-functional reference implementations or problems to evaluate them at scale.

% Table generation script:
% /home/azureuser/code/Users/maximilian.schambach/tabgemma-eval.ipynb
\input{tables/eval_table}

% \begin{table}[t]
% \centering 
% \footnotesize
% \setlength{\tabcolsep}{3pt} 
% % tighter columns 
% \sisetup{ table-number-alignment = center, round-mode = places, round-precision = 1, detect-weight = true, detect-family = true } 
% \caption{Comparison of tabular models on CARTE, TextTab, and TabArenaLight.\tassilo{can we divide the approaches based on architecture, e.g., TabPFN, Tree, etc.}}
% \label{tab:tabular-benchmarks} 
% \begin{adjustbox}{max width=\textwidth} 
% \input{tables/eval_table}
% \end{adjustbox} 
% \end{table}

\custompar{Results}
The main results are summarized in Table~\ref{tab:tabular-benchmarks}. 
Here, we report TabGemma results using $k=128$ retrieved context examples by default. To confirm the efficacy of our approach, we also evaluated an off-the-shelf Gemma 3 model on CARTE. In the absence of prompt engineering, performance is poor, \num{4.6}\% accuracy and \num{-98.4} $R^2$, and was hence excluded.
On the semantically rich CARTE and TextTab benchmarks, TabGemma matches or surpasses state-of-the-art baselines in classification performance. 
To the best of our knowledge, this is the first time an LLM-based approach outperforms extensively tuned baselines, as well as AutoGluon, which stacks a multitude of per-dataset tuned predictors. Note that evaluation is done at the full scale of the available datasets. 

However, TabGemma lags behind on regression and on TabArena generally. However, on CARTE, regression performance of TabGemma still surpasses extensively tuned LGBM, as well as tuned and ensembled RealMLP. 
The poor regression performance on TextTab is surprising and needs to be investigated more closely in the future.
As TabArena focuses on conventional, numerically dominated tasks, this underscores current limitations of language models in this regime and highlights further potential in tokenization. 
While weaker regression performance is expected, our rank-based analysis further indicates that retrieval and long-context handling degrade as the number of rows and columns increases. 
We investigate this in more detail in the Appendix~\ref{app:add-results}.

\section{Discussion and Conclusion}
We presented TabGemma: an LLM-based in-context learner that combines improved numeric representation, retrieval, and continued pretraining over a large corpus of real-world tables. On semantically rich benchmaks, TabGemma delivers strong classification in both low- and high-data regimes, outperforming extensively tuned baselines and AutoML solutions, while revealing gaps on regression and on wide or very large tables due to context and retrieval constraints.
While shining in the few-shot regime, to the best of our knowledge, this is the first time an LLM-based tabular predictor outperforms baselines also at full dataset evaluation.

There are a several limitations of our current approach opening a multitude of future research directions: first, we observe that TabGemma is very effective at classification but underperforms in regression as well as tasks with numeric-heavy features. This motivates further research into numerics-adapted tokenization. Second, performance on very large and wide tables degredates. Further research into more compact tokenization schemes as well as even more efficient retrieval approaches may mitigate this. And last, note that the proposed table serialization and autoregressive modeling breaks the natural column and row order permutation equivariance inherent to many tabular prediction problems. The effect of this on tabular language modeling needs to be investigated more closely openining potential gains when ensembling over permuted inputs.

\section*{Acknowledgements}
We would like to thank Johannes Hoehne, Johannes Hoffart, and Markus Kohler for their insightful comments and suggestions throughout the development of this work. We thank Thassilo Klein for providing valuable feedback on the draft of this contribution.

% Our key takeaways are: (1) LLMs can be effective tabular in-context learners; semantics transfer especially in data-scarce settings. (2) Numeric encoding is the main bottleneck for regression. (3) Long-context handling and retrieval limit scalability with many rows/columns.

% Future directions: While showing initially promising results, our approach shall be evaluated and investigated in more depth in the future. In particular, its performance on very large or wide tables as well as regression tasks. Also, note that the proposed table serialization and autoregressive modeling breaks the natural column and row order permutation equivariance inherent to many tabular prediction problems. The effect of this on tabular language modeling needs to be investigated more closely.

{
\small

\bibliography{refs}
\bibliographystyle{plainnat}

}

\clearpage
\newpage

\appendix

\section{Additional results}\label{app:add-results}

\subsection{Critical difference diagrams}
\begin{figure}[h]
\centering
\begin{subfigure}[t]{\textwidth}
    \includegraphics[width=0.495\linewidth]{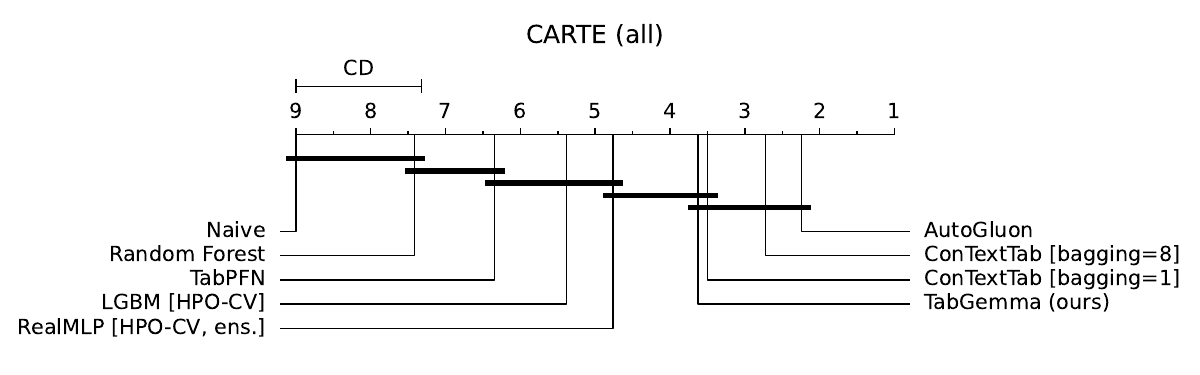} \hfill\\
    \includegraphics[width=0.495\linewidth]{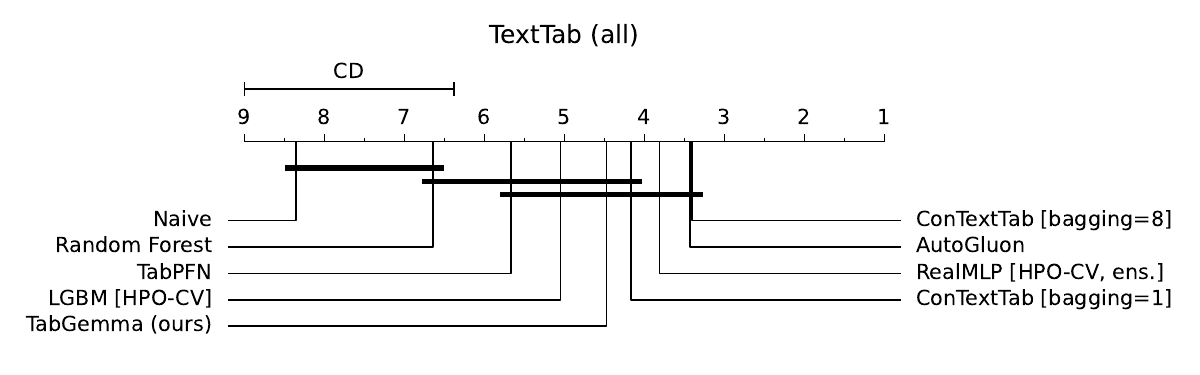} \hfill
    \includegraphics[width=0.495\linewidth]{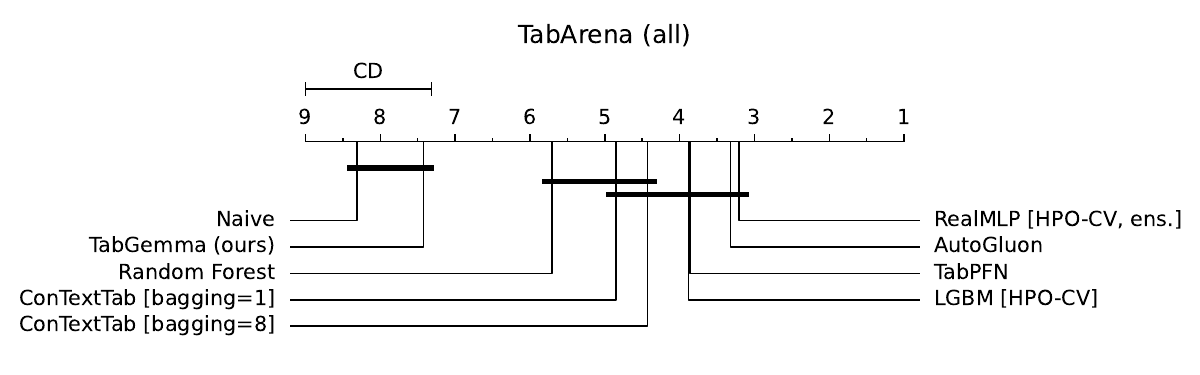} 
    \caption{All tasks.}
\end{subfigure}\vspace{3mm}
\begin{subfigure}[t]{\textwidth}\centering
    \includegraphics[width=0.495\linewidth]{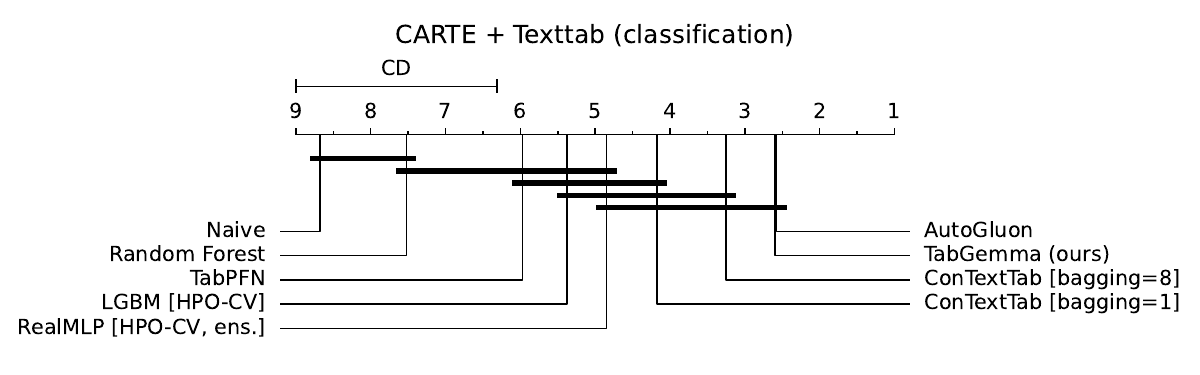} \hfill
    \includegraphics[width=0.495\linewidth]{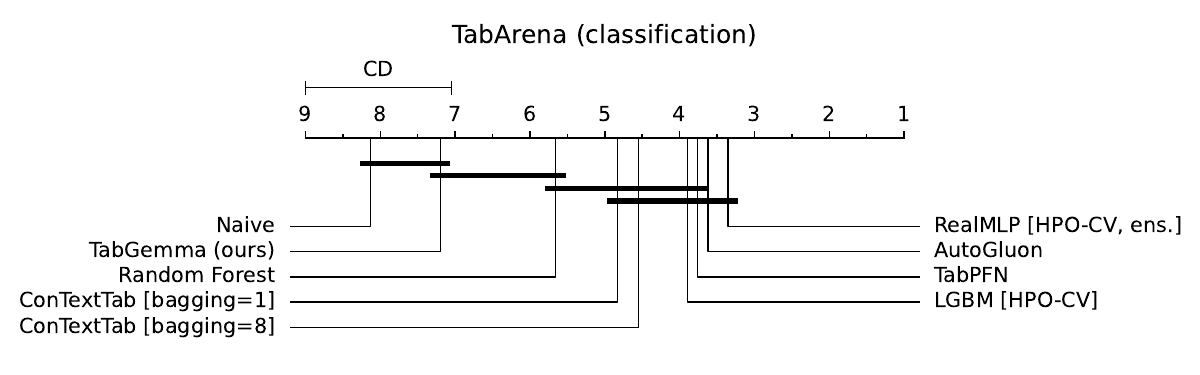} 
    \caption{Classification tasks only.}
\end{subfigure}\vspace{3mm}
\begin{subfigure}[t]{\textwidth}\centering
    \includegraphics[width=0.495\linewidth]{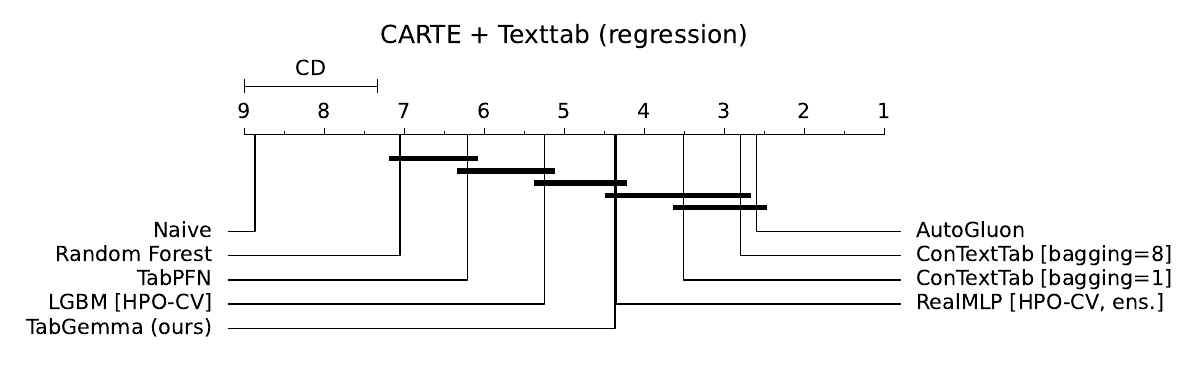} \hfill
    \includegraphics[width=0.495\linewidth]{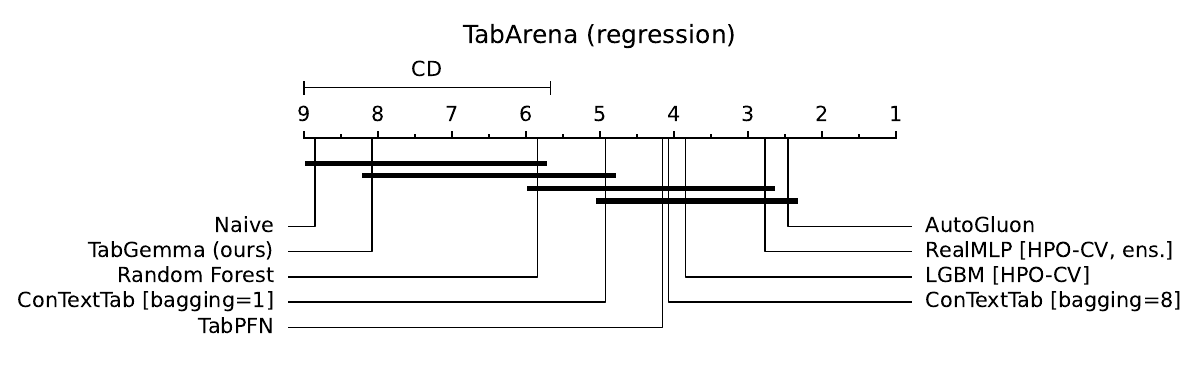}
    \caption{Regression tasks only.}
\end{subfigure}
\caption{Critical difference diagrams on the investigated benchmarks, including all datasets (``all'') as well as classification- and regression-only subsets. Note that, due to the limited support of classification tasks on CARTE and TextTab, evaluation was performed over the union of tasks in this case.}
\label{fig:cd-diagrans}
\end{figure}

We provide critical difference (CD) diagrams in Figure~\ref{fig:cd-diagrans}. To this end, we use the \texttt{autorank} package\footnote{\href{https://github.com/sherbold/autorank}{github.com/sherbold/autorank}}. We calculate CD-diagrams on the full benchmark as well as on classification- and regression-only subsets. Note that, since the support set for classification tasks on CARTE and TextTab were too small to calculate the CD, we show a joined evaluation on the two instead.

We observe that TabGemma performs SOTA on semantic classification tasks within CARTE and TextTab, on par with AutoGluon. While its regression performance lags behind here, it is not statistically significantly worse then other state-of-the-art approaches such as ConTextTab or RealMLP.

On Tabarena, the performance of TabGemma is noticeably worse, statistically on par with a non-tuned Random Forest predictor. This shortcoming on non-semantic tasks opens future research directions.

\clearpage\newpage
\subsection{Sample efficiency and few-shot domain}
\begin{figure}[hb]
\centering
\begin{subfigure}[t]{0.48\textwidth}
    \includegraphics[width=\linewidth]{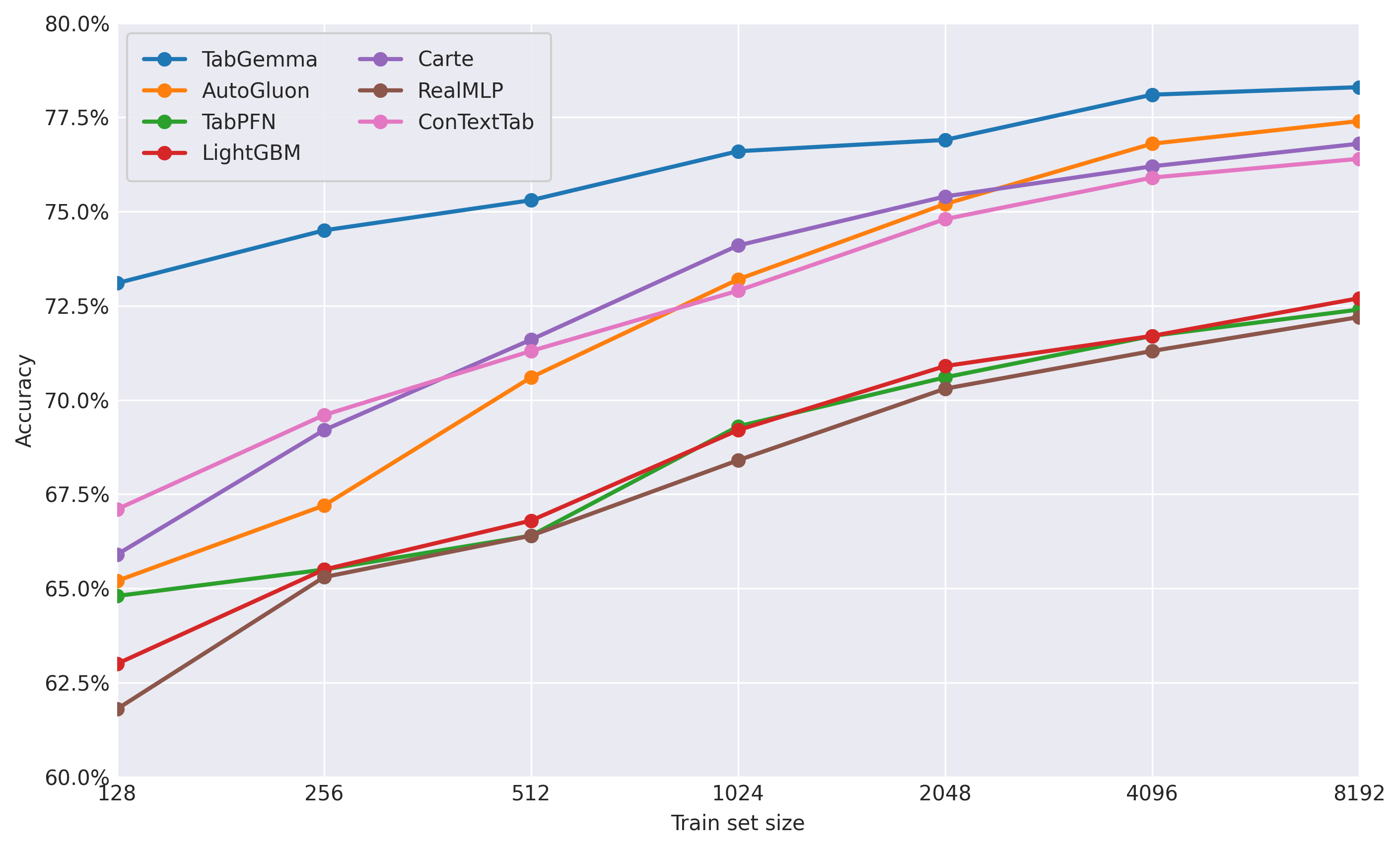} 
    \caption{Average Accuracy on classification.}
\end{subfigure}\hfill
\begin{subfigure}[t]{0.48\textwidth}
    \includegraphics[width=\linewidth]{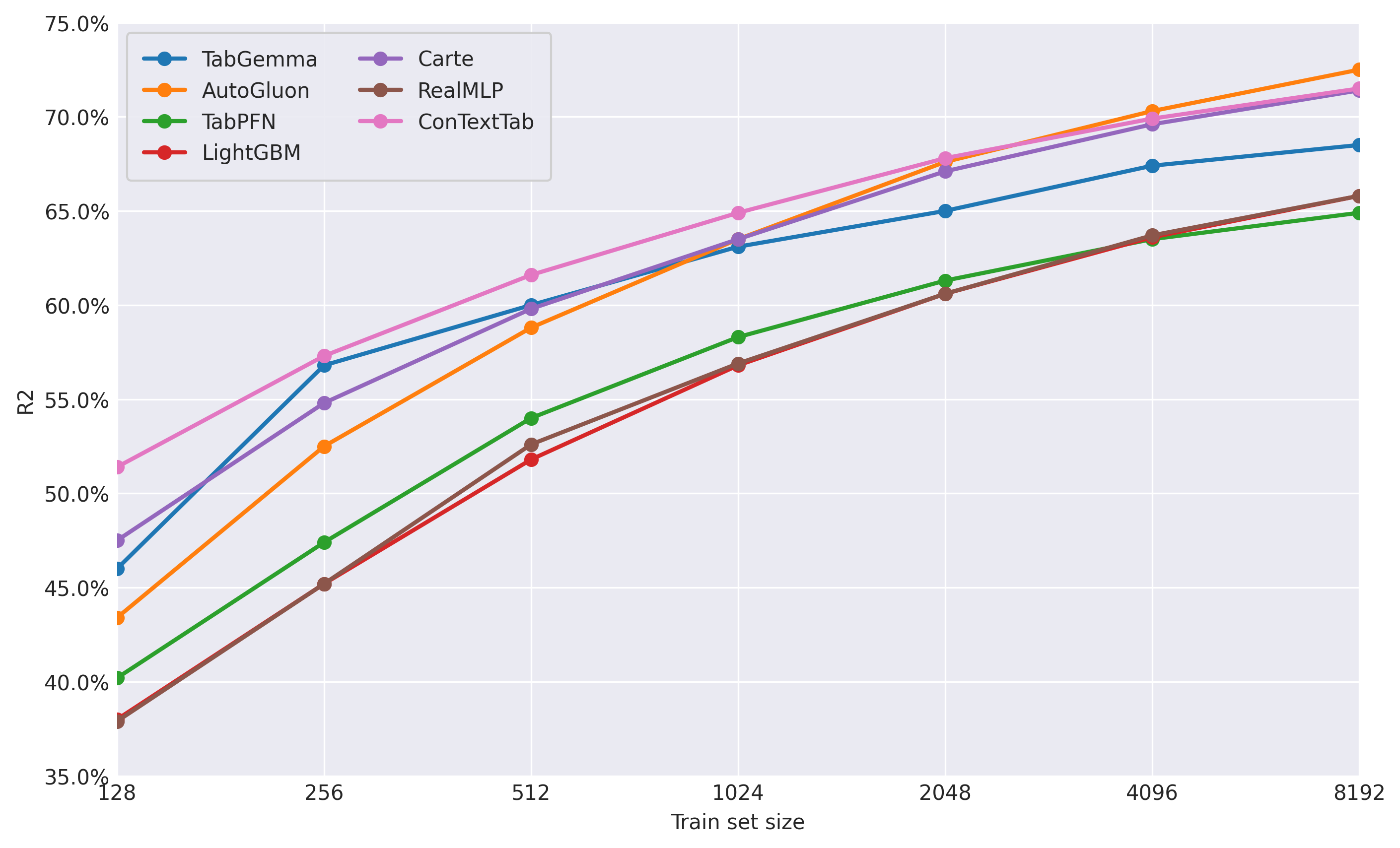}
    \caption{Average R2 on regression.}
\end{subfigure}
\caption{Results on the CARTE benchmark for varying train subsets.}
\label{fig:subset}
\end{figure}

Figure \ref{fig:subset} examines performance as a function of available training data. 
To this end, using the CARTE benchmark datasets, we subsample each training set to {128, 256, …, 8192} rows and evaluate each model under the same subset: baselines are trained on that subset, whereas TabGemma receives the same subset as its retrieval pool and in-context exemplars. 
In low-data regimes, TabGemma is highly competitive on classification and surpasses the baselines by a notable margin. 
This observation is in line with previous LLM-based results such as TabuLa~\cite{tabula8b} but stretches to much larger shot examples, whereas Tabula reported results only in the very few-shot domain of up to 32 samples.
In the few-shot domain and particularly for highly semantic tasks, LLMs likely benefit most from their world knowledge obtained during their extensive pretraining.

For regression, TabGemma keeps pace up to roughly 1024 training examples but lags behind AutoGluon and CARTE once more data are available.

\subsection{Scalability with context size}\label{sec:context}

\begin{figure}[hb]
\centering 
\begin{subfigure}[t]{0.48\textwidth} 
\includegraphics[width=\linewidth]{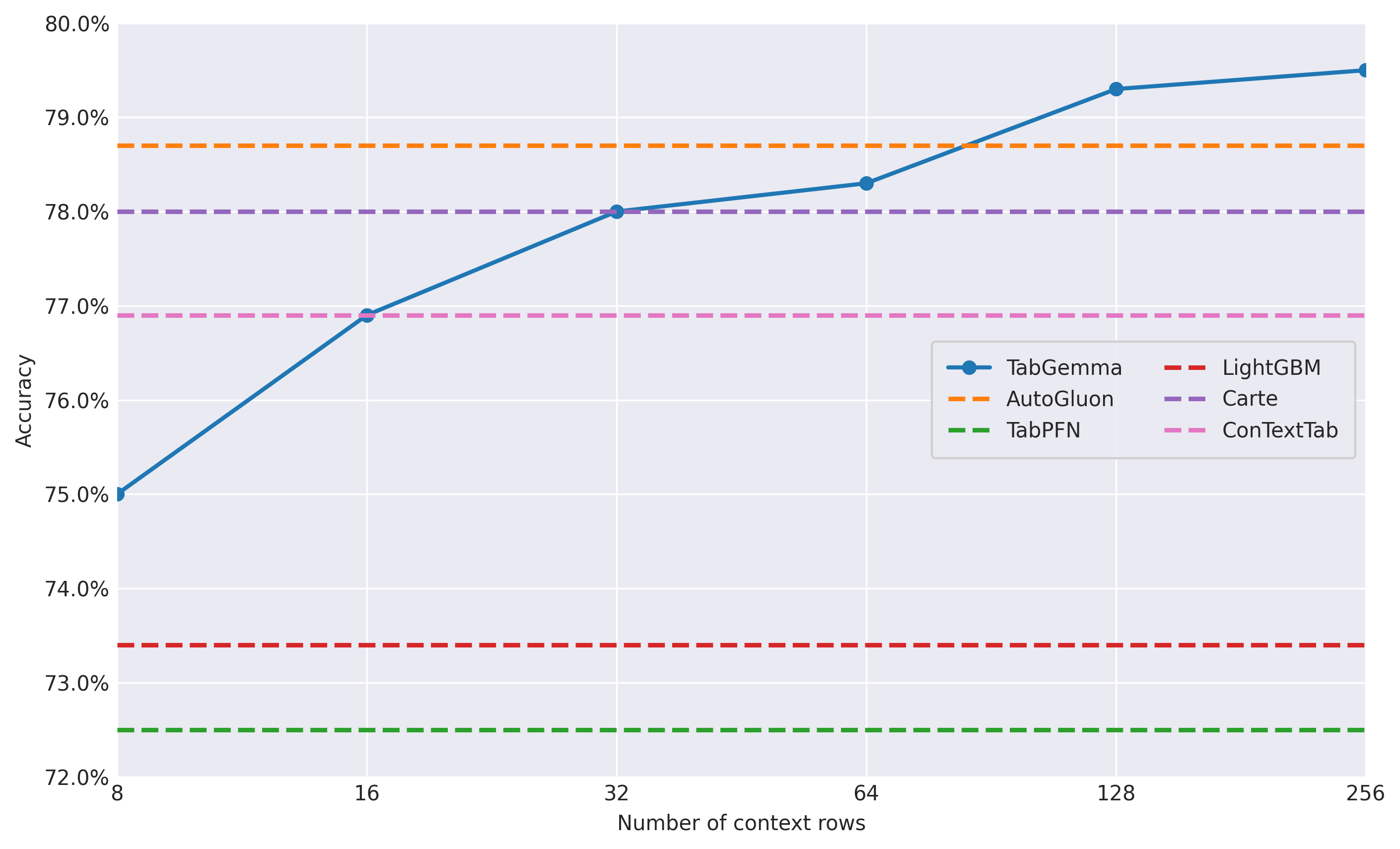} 
\caption{Average Accuracy on classification.} 
\label{fig:a} 
\end{subfigure}\hfill
\begin{subfigure}[t]{0.48\textwidth}
\includegraphics[width=\linewidth]{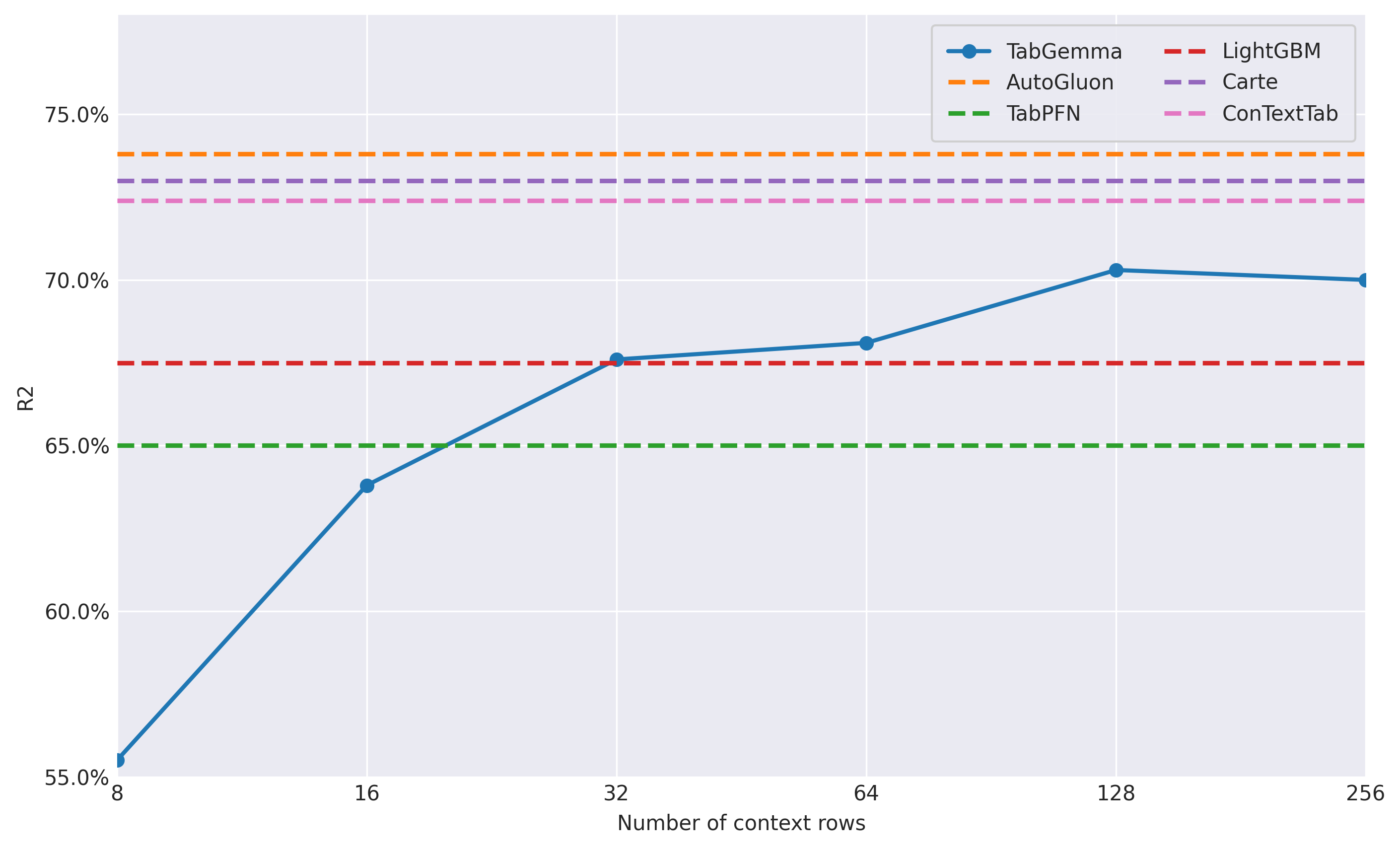}
\caption{Average $R^2$ on regression.}
\label{fig:b}
\end{subfigure}
\caption{Results on the CARTE benchmark for varying context rows.} 
\label{fig:context} 
\end{figure}

Next, we studiey how TabGemma scales as we increase the number of context rows, i.e., the number of retrieved training examples included in the prompt. 
The results are depicted in Figure~\ref{fig:context}. 
Dashed lines denote baselines trained on the full training split (or, for ICL-style baselines, using the full training pool as their retrieval source). 
Solid lines denote TabGemma with varying number of $k$ context rows and no gradient updates. 
As $k$ grows, TabGemma improves monotonically and establishes a new state-of-the-art on classification within CARTE. 
On regression, however, AutoGluon, CARTE and ConTextTab remain ahead, highlighting a current limitation of our numeric handling for continuous targets. 
Nevertheless, TabGemma still outperforms an extensivly tuned and ensembled LGBM predictor.

\section{Limitations across task scale and dimensionality}
\label{sec:rank}
\begin{figure}[hb]
\centering 
\begin{subfigure}[t]{\textwidth} 
\includegraphics[width=0.475\linewidth]{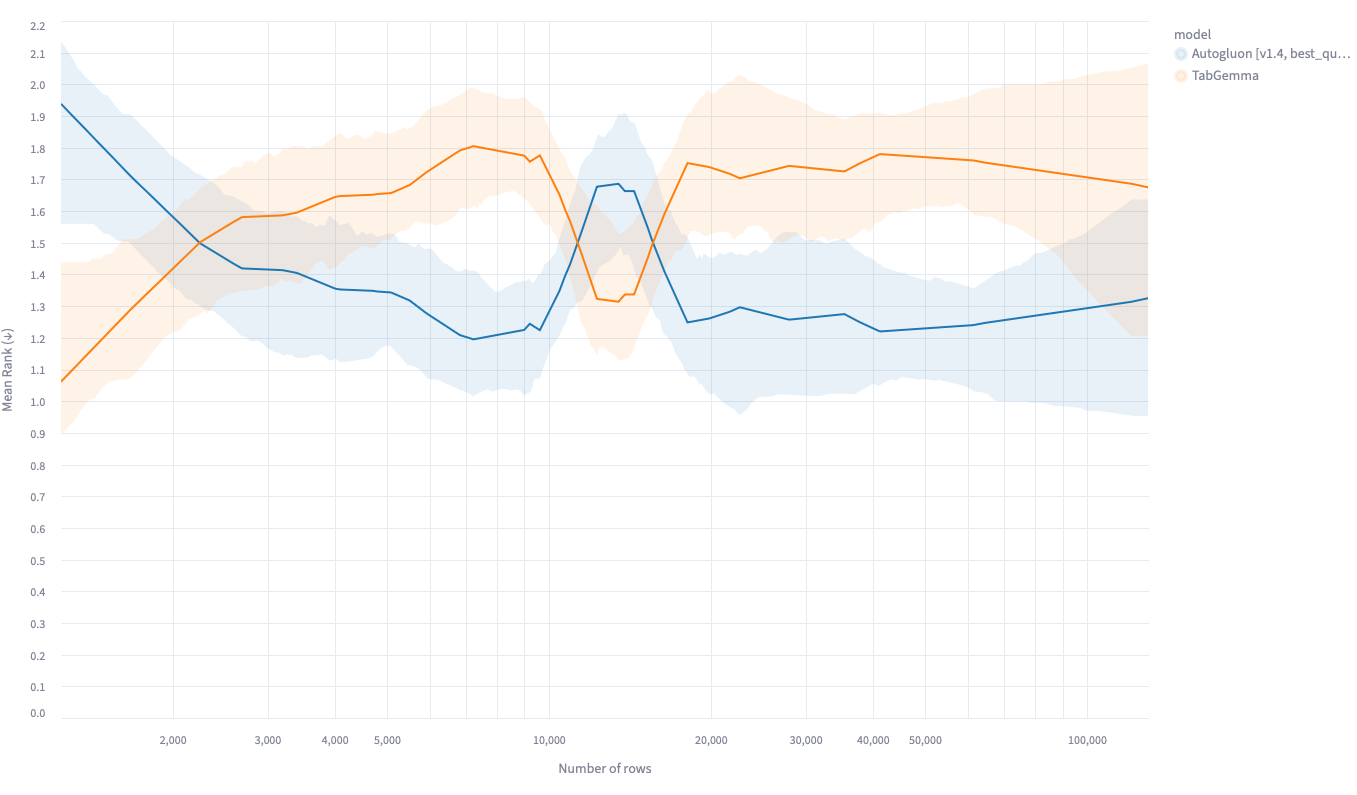} \hfill
\includegraphics[width=0.475\linewidth]{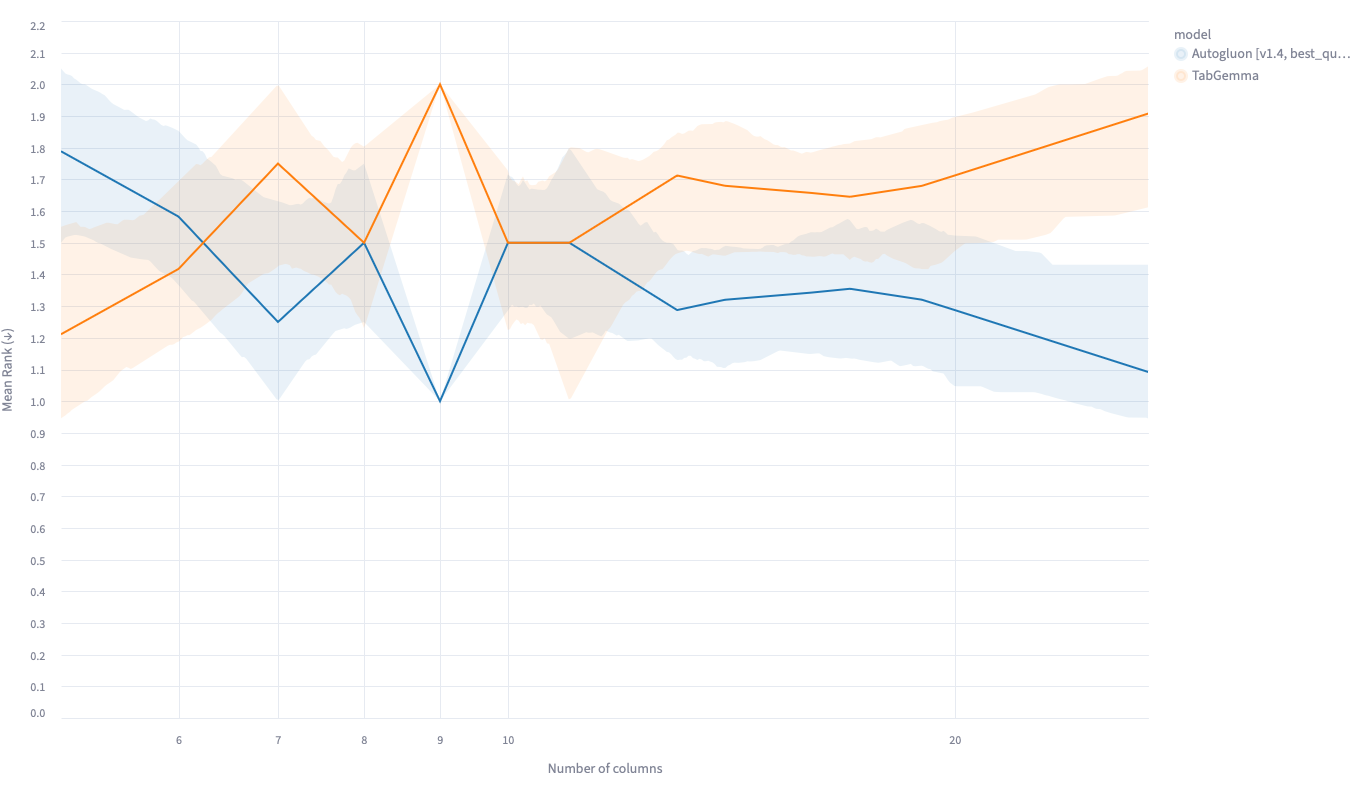} 
\caption{CARTE.} 
\end{subfigure}\vspace{2mm}
\begin{subfigure}[t]{\textwidth} 
\includegraphics[width=0.475\linewidth]{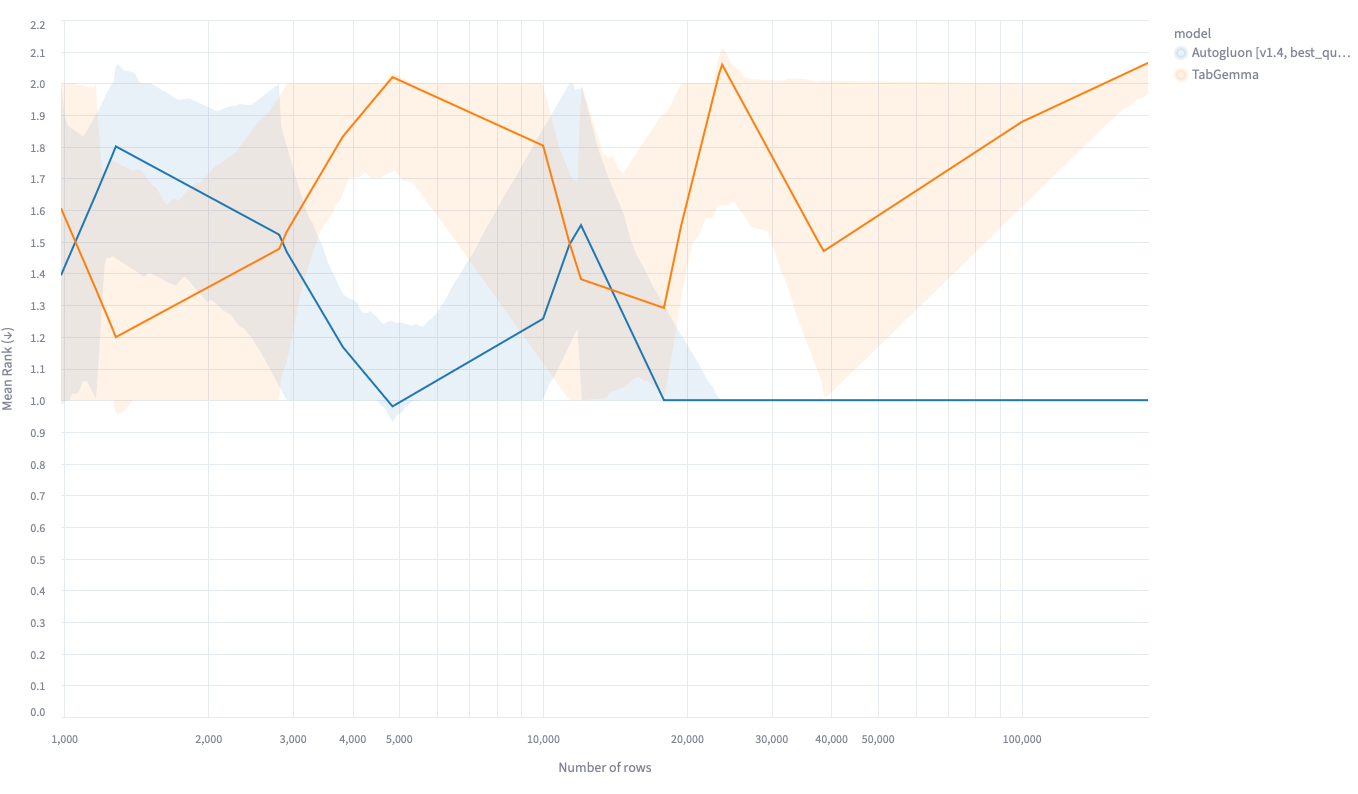} \hfill
\includegraphics[width=0.475\linewidth]{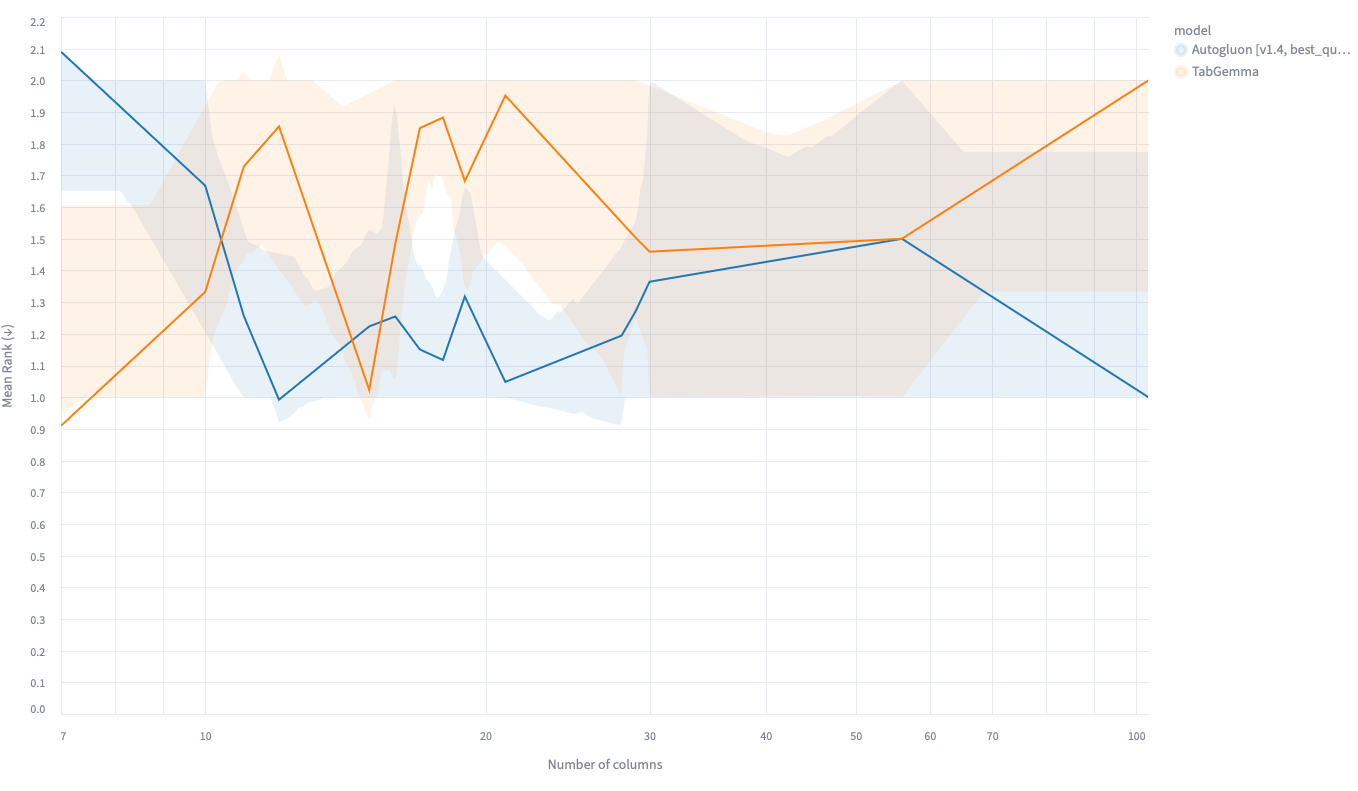} 
\caption{TextTab.} 
\label{fig:d} 
\end{subfigure}\vspace{2mm}
\begin{subfigure}[t]{\textwidth} 
\includegraphics[width=0.475\linewidth]{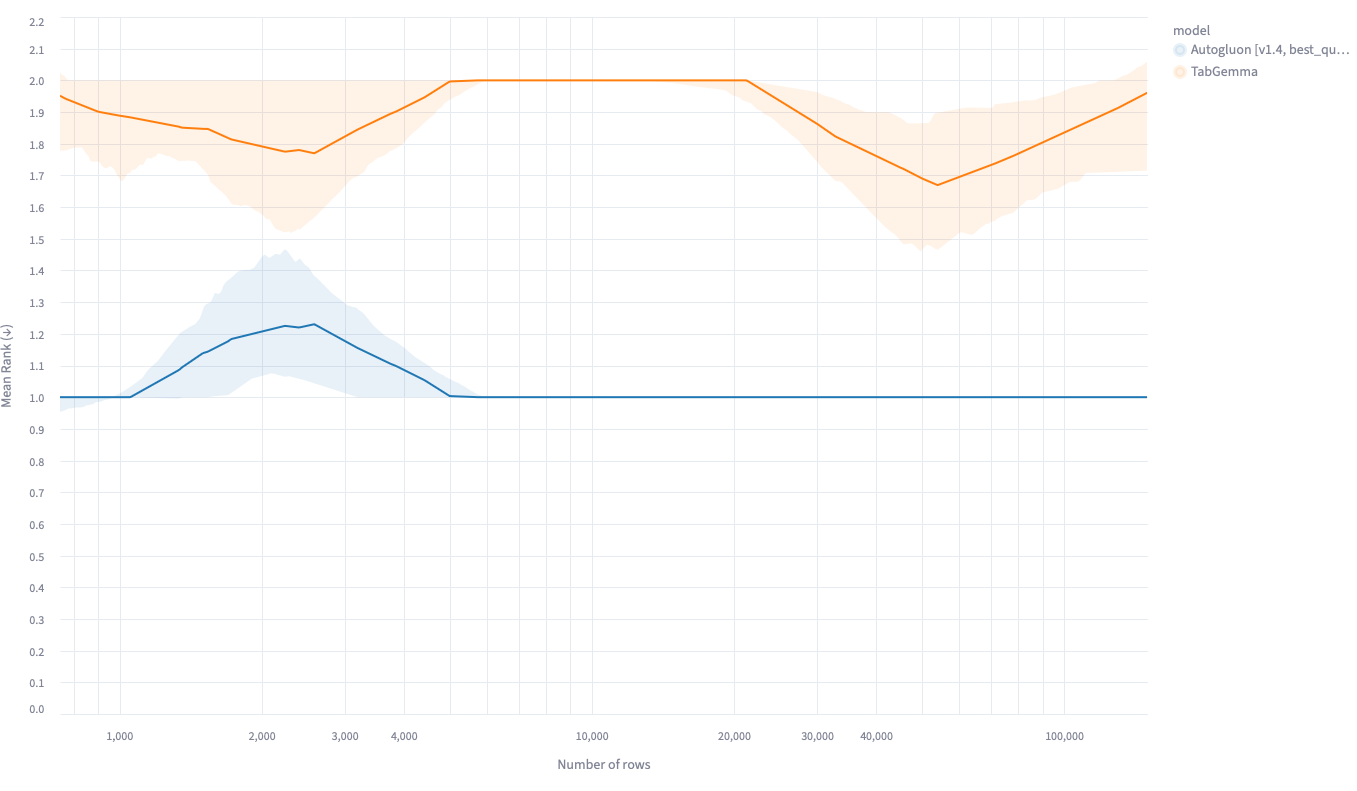} \hfill
\includegraphics[width=0.475\linewidth]{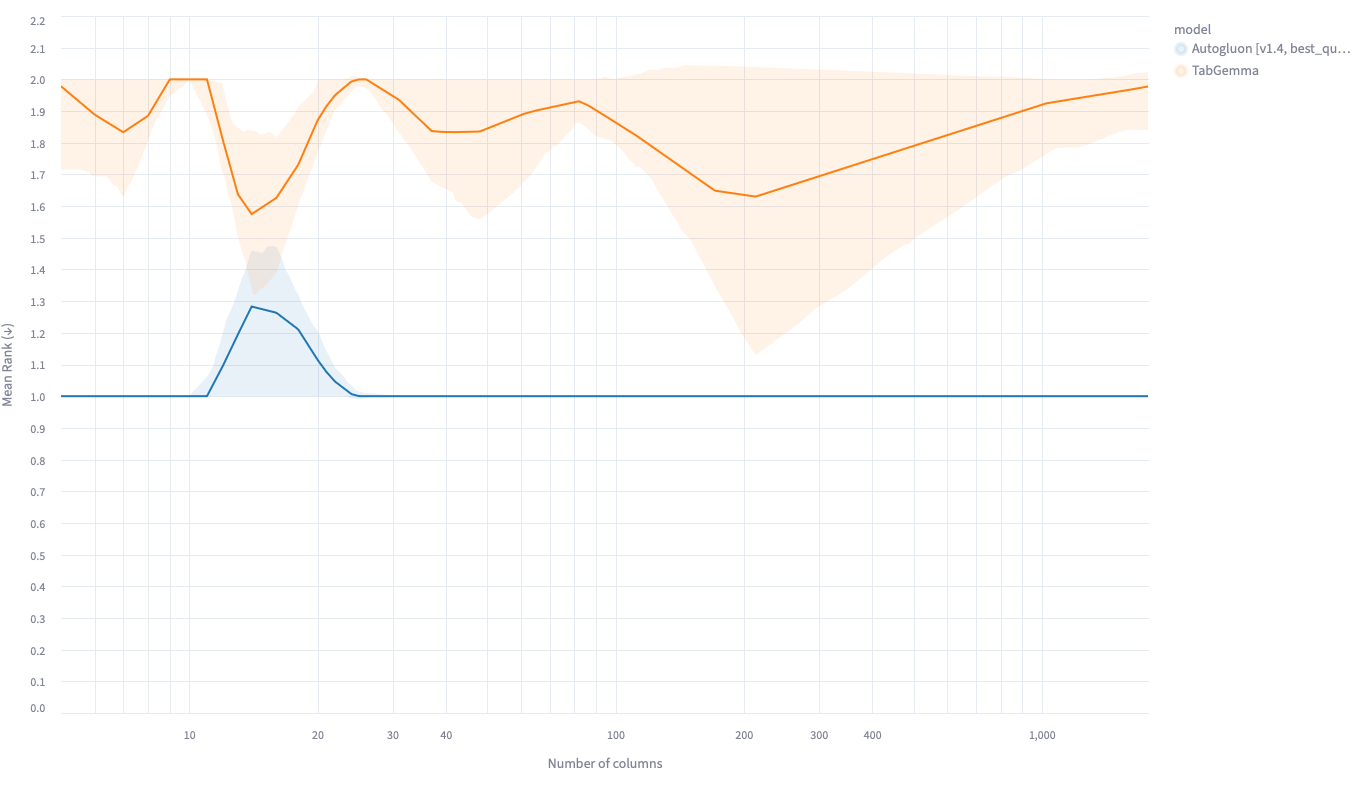} 
\caption{TabArena.} 
\end{subfigure} 
\caption{Dependence of mean rank (lower is better) of AutoGluon (blue) vs. TabGemma (orange) on the number of rows (left) and number of columns (right) across different evaluated benchmarks.} 
\label{fig:rank} 
\end{figure}

Figure \ref{fig:rank} reports mean rank (lower is better) for TabGemma versus AutoGluon across CARTE, TextTab, and TabArena, stratified by training-set size and the number of columns. 
AutoGluon consistently leads in regimes with more training rows and wider tables, and the gap widens as scale increases. 
These trends suggest two main bottlenecks for TabGemma: (1) retrieval precision declines as the candidate pool grows, reducing the quality of in-context examples; and (2) the 128k-token context window limits how many rows and columns can be represented without lossy compression, weakening long-context modeling.

\clearpage\newpage
\section{Evaluation datasets and baselines}

\subsection{Datasets}
As previously discussed, we evaluate all methods on the CARTE, TextTab, and TabArena-Lite benchmark covering 123 tasks in total.
We use the CARTE data from the official reference implementation\footnote{\href{https://github.com/soda-inria/carte}{github.com/soda-inria/carte}}. CARTE contains 51 tasks in total, 11 exclusively binary classification tasks and 40 regression tasks. We create custom 80:20 train target splits using the target column for stratification.
TextTab contains 21 tasks in total, 9 mostly binary classification tasks and 12 regression tasks. We use all tasks from the original publication~\cite{texttabbench} as well as the ``extra datasets'' given in the reference implementation\footnote{\href{https://github.com/mrazmartin/TextTabBench}{github.com/mrazmartin/TextTabBench}}. Again, we create custom stratified 80:20 train test splits.
TabArena contains 51 tasks in total, 38 mostly binary classification tasks and 13 regression tasks. We use the splits as defined in the official OpenML release\footnote{\href{https://www.openml.org/search?type=study&study_type=task&id=457}{openml.org/search?type=study\&study\_type=task\&id=457}} of the benchmark's lite variant covering the first fold only.

\begin{figure}
\centering
\begin{subfigure}[t]{0.475\textwidth}
    \includegraphics[width=\linewidth]{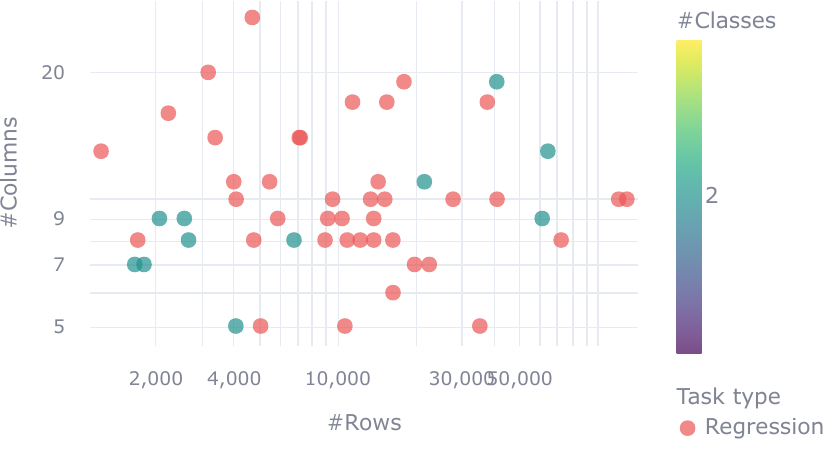} 
    \caption{CARTE.}
\end{subfigure}\\[3mm]
\begin{subfigure}[t]{0.475\textwidth}
    \includegraphics[width=\linewidth]{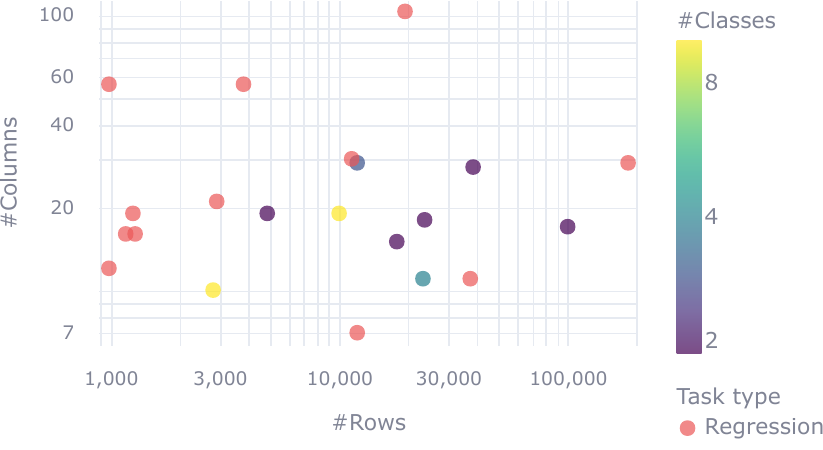} 
    \caption{TextTab.}
\end{subfigure}\hfill
\begin{subfigure}[t]{0.475\textwidth}
    \includegraphics[width=\linewidth]{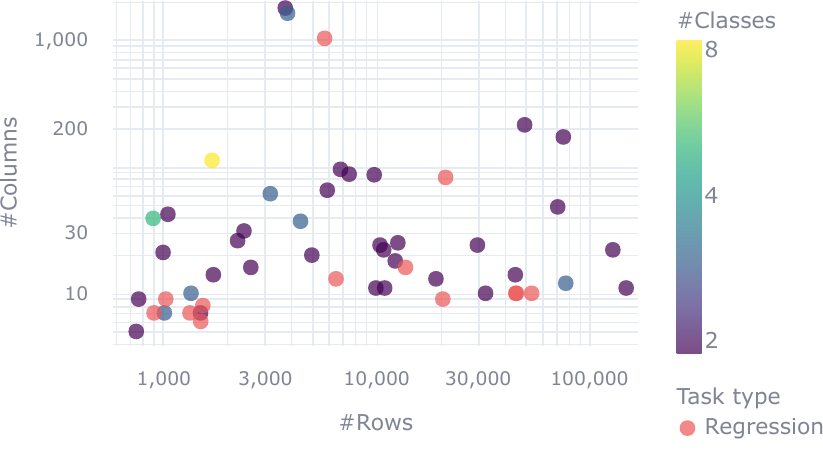} 
    \caption{TabArena.}
\end{subfigure}
\caption{Dataset statistics of the evaluated benchmarks.}
\label{fig:dataset-stats}
\end{figure}

The task size distribution of all evaluated benchmarks is depicted in Figure~\ref{fig:dataset-stats}.

\subsection{Baselines}\label{app:baselines}

Throughout, we follow the evaluation protocol of~\cite{contexttab}. That is, we use a AutoGluon-based standardized feature encoder for all baselines that do not provide a custom one. In particular, the encoder natively handles categorical data, free text (via conventional NLP features), as well as datetime encoding. In particular, we evaluate the following model versions.

\custompar{Pytabkit models} 
We use the \texttt{pytabkit}~\cite{realmlp} for evaluating RealMLP, and LightGBM. 
We evaluate LightGBM with ensembled hyperparameter optimization across 5-fold inner cross-validation (HPO-CV). For RealMLP, we do the same but combine it with the recently introduced learned ensemble, further pushing its performance (HPO-CV, Ens.).
For the HPO variants, we use the recently added \texttt{tabarena} search spaces proposed in~\cite{tabarena}.

\custompar{TabPFN}
We use the model from the official \texttt{tabpfn} package at version 2.1.0 with the
\texttt{tabpfn-extensions} package version 0.1.0.
For datasets larger than the native 10\,k limit of TabPFN, we sample a random 10\,k subset of the training split. For datasets with more than the 500 feature limit, we select a random subsample of 500 features. 

\custompar{ConTextTab}
We evaluate ConTextTab v1.0.1 using the reference implementation and checkpoint\footnote{\href{https://github.com/SAP-samples/contexttab}{github.com/SAP-samples/contexttab}}. We set a context size of 8k samples and evaluate variants without and with 8-fold bagging.

\custompar{CARTE}
We use the model provided in the official \texttt{carte-ai} package with version 0.0.26. 
We use \texttt{CARTEClassifier} and \texttt{CARTERegressor} with default parameters for classification and regression tasks, respectively.

\custompar{Sklearn models}
We evaluate the Random Forest and Naive baseline models from \texttt{scikit-learn}~\citep{sklearn}, combining them with the default preprocessor as outlined above. Evaluation is performed using \texttt{scikit-learn} v1.5.2.

For the naive predictor, we use the \texttt{DummyClassifier} and \texttt{DummyRegressor} to predict the most frequent, respectively mean value of the train splits as the naive majority baseline.

For the random forest predictor, we use the \texttt{RandomForestClassifier} and \texttt{RandomForestRegressor} for classification and regression tasks, respectively, using default hyperparameters. The model handles missing values natively.

\custompar{AutoGluon}
We evaluate using AutoGluon v1.4 with its native feature encoder. We use the \texttt{best\_quality} preset with a per-dataset time limit of 4\,h. (We have found the ``extreme'' preset to only yield slightly better results at the expense of much higher compute requirements).

\end{document}

%% file: tables/eval_table.tex
\begin{table}\centering\footnotesize
\caption{Evalation results on the investitated benchmarks sorted by classification performance on CARTE. We report accuracy (Acc) and (soft-clipped) R2 for classification and regression, respectively.}
\label{tab:tabular-benchmarks}
\begin{tabular}{lrrrrrrrr}
\toprule
 & \multicolumn{2}{c}{\textbf{All}} & \multicolumn{2}{c}{\textbf{CARTE}} & \multicolumn{2}{c}{\textbf{TextTab}} & \multicolumn{2}{c}{\textbf{TabArena}} \\
  \cmidrule(lr){2-3} \cmidrule(lr){4-5} \cmidrule(lr){6-7} \cmidrule(lr){8-9}
Model & \multicolumn{1}{c}{\textbf{Acc}} & \multicolumn{1}{c}{\textbf{R$^\textrm{2}$}}  & \multicolumn{1}{c}{\textbf{Acc}} & \multicolumn{1}{c}{\textbf{R$^\textrm{2}$}}  & \multicolumn{1}{c}{\textbf{Acc}} & \multicolumn{1}{c}{\textbf{R$^\textrm{2}$}}  & \multicolumn{1}{c}{\textbf{Acc}} & \multicolumn{1}{c}{\textbf{R$^\textrm{2}$}}  \\
\midrule
\bfseries TabGemma (ours) & 83.6 & 60.7 & \bfseries 79.3 & 70.3 & 84.1 & 31.6 & 84.8 & 57.8 \\
AutoGluon & \bfseries 85.9 & \bfseries 70.5 & 78.9 & \bfseries 73.4 & 83.9 & 51.8 & 88.5 & 78.9 \\
ConTextTab [bagging=8] & 85.0 & 70.3 & 76.9 & 72.4 & \bfseries 84.3 & \bfseries 55.0 & 87.6 & 77.9 \\
ConTextTab [bagging=1] & 84.9 & 69.6 & 76.4 & 71.5 & 84.1 & 54.4 & 87.5 & 77.6 \\
RealMLP [HPO-CV, ens.] & 84.6 & 67.6 & 73.6 & 68.2 & 81.9 & 52.2 & \bfseries 88.5 & \bfseries 79.8 \\
LGBM [HPO-CV] & 84.4 & 66.3 & 73.4 & 67.5 & 81.5 & 48.6 & 88.2 & 78.9 \\
TabPFN & 83.2 & 63.2 & 72.3 & 65.0 & 81.6 & 41.9 & 86.7 & 77.0 \\
Random Forest & 83.3 & 62.5 & 71.5 & 63.3 & 79.8 & 45.3 & 87.6 & 76.3 \\
Naive & 70.1 & -3.5 & 53.0 & -1.8 & 70.4 & -5.0 & 75.0 & -7.3 \\
%Gemma 3 12B & \NAN & \NAN & 4.6 & -98.4 & \NAN & \NAN & \NAN & \NAN \\
\bottomrule
\end{tabular}
\end{table}